\pdfoutput=1

\documentclass[11pt]{article}
\usepackage[table,xcdraw]{xcolor} 
\usepackage[final]{acl}

\usepackage{times}
\usepackage{latexsym}

\usepackage{amssymb}
\usepackage{comment}
\usepackage{arydshln}
\usepackage{longtable}
\usepackage{multirow}
\usepackage{graphicx}
\usepackage{url}

\usepackage{booktabs}
\usepackage{tikz}
\usepackage{tabularray}
\usepackage[T1]{fontenc}
\usepackage{multirow}

\definecolor{coc green}{RGB}{241, 251, 224} 
\definecolor{pastelgreen}{RGB}{229, 245, 194} 

\usepackage{arydshln}

\usepackage[T1]{fontenc}

\usepackage[utf8]{inputenc}

\usepackage{microtype}

\usepackage{inconsolata}

\usepackage{graphicx}
\usepackage{xcolor} 
\usepackage{booktabs}
\usepackage{tikz}
\usepackage[table]{xcolor}
\usepackage{tabularray}
\usepackage[T1]{fontenc}
\usepackage{multirow}

\usepackage[table,xcdraw]{xcolor} 

\definecolor{coc green}{RGB}{241, 251, 224} 
\definecolor{pastelgreen}{RGB}{229, 245, 194} 
\title{Retrieval Augmented Generation based Large Language Models for Causality Mining}

 \author{Thushara Manjari Naduvilakandy, Hyeju Jang, Mohammad Al Hasan\thanks{\quad Corresponding author.} \\
 Department of Computer Science
          \\ Indiana University Indianapolis\\
           \texttt{tnaduvil@iu.edu},  \texttt{hyejuj@iu.edu},  \texttt{alhasan@iu.edu}}

\begin{document}
\maketitle
\begingroup\renewcommand\thefootnote{}\footnotetext{
Code: \url{https://github.com/thusharamanjari/RAG_LLM_CausalityMining}
}\endgroup

\begin{abstract}

Causality detection and mining are important tasks in information retrieval due
to their enormous use in information extraction, and knowledge graph
construction. To solve these tasks, in existing literature there exist several
solutions---both  unsupervised and supervised. However, the unsupervised methods
suffer from poor performance and they often require significant human
intervention for causal rule selection, leading to poor generalization across
different domains. On the other hand, supervised methods suffer from the lack of
large training datasets. Recently, large language models (LLMs) with effective
prompt engineering are found to be effective to overcome the issue of
unavailability of large training dataset. Yet, in existing literature, there
does not exist comprehensive works on causality detection and mining using LLM
prompting. In this paper, we present several retrieval-augmented generation
(RAG) based dynamic prompting schemes to enhance LLM performance in causality
detection and extraction tasks. Extensive experiments over three datasets and
five LLMs validate the superiority of our proposed RAG-based dynamic prompting
over other static prompting schemes.
\end{abstract}
\section{Introduction}

The expression of causality, along with causal reasoning, is a defining
characteristic of human cognition~\cite{Goddu24}, which makes natural language
text a trove of causal knowledge. Extracting this knowledge from text is a
significant task due to its widespread applications in natural language
processing (NLP), including question answering~\cite{khoo1998automatic,
girju2003automatic}, event
prediction~\cite{radinsky2012learning,silverstein2000scalable} and  medical text
mining~\cite{hashimoto2014toward,riaz2010another}. The importance of causality
extraction task has intensified even further with the proliferation of AI
Chatbots and conversational AI agents, which rely on synthesized causal knowledge.
Therefore, sophisticated and scalable models are essential to
accurately extract causal knowledge from various textual sources, ranging from
scientific documents to news articles.

Causality in natural language text refers to the relationship between two entities in a sentence, where the occurrence of one entity (the cause) leads to the occurrence of the other (the effect). There are two prominent research tasks related to causality: causality detection~\cite{blanco2008causal, hidey2016identifying, kayesh2019event} and causality extraction~\cite{dasgupta2018automatic, li2021causality, kabir2023deep}. The first task detects whether a given sentence is causal or not, which is typically framed as a binary classification task. The second task labels the cause and the effect phrases within a given sentence, given that it is causal.
This task is often formulated as a sequence labeling task, where each token in the sentence is labeled as part of a cause phrase, effect phrase, or neither. 

In traditional machine learning, many unsupervised~\cite{kaplan1991knowledge,
garcia1997coatis, girju2002text} and supervised~\cite{li2021causality,
dasgupta2018automatic, kabir2023deep}  methods are proposed for causality
mining. Among these, the unsupervised methods require significant human
intervention, and the supervised methods suffer from the lack of large training
datasets. With the emergence of pre-trained large language models (LLMs) such as
GPT~\cite{brown2020language, achiam2023gpt}, Llama~\cite{touvron2023llama},
Gemini~\cite{anil2023gemini}, and Mixtral~\cite{jiang2023mistral}, many
supervised NLP tasks are now being solved using LLMs, primarily due to their
ability to produce superior results only with minimal supervision. This is
exciting news for causality mining, as 
supervised models for causality mining, trained on datasets from one domain, often perform poorly
on datasets from different domains. Thus, LLMs with few-shot learning
capabilities could be an attractive approach for causality mining. 

Performance improvements of LLMs in specific tasks like causality mining can be
achieved through fine-tuning, prompt engineering, and Retrieval-Augmented
Generation (RAG) approaches. Fine-tuning methods train weights of a few terminal
layers of the LLMs, prompt engineering selects effective in-context examples for
few-shot training of LLMs, and RAG improves LLM outputs by generating prompt
dynamically by leveraging relevant knowledge from external sources. For
causality mining using LLMs, \citet{jin2023can} proposed to improve LLM
performance in causal inference by fine-tuning, but their results show that
fine-tuned models fail to generalize, yielding poorer results on
out-of-distribution data. Besides, fine-tuning large language models requires
significant resources and diverse training data as they are not good for
incorporating dynamic knowledge based on varying scenarios. These limitations of
fine-tuning open the frontier of research in prompt engineering and RAG to
improve  LLM performance. 


In this paper, we propose two novel RAG (Retrieval-Augmented Generation)
approaches namely \textbf{Pattern RAG} and \textbf{$k$NN+Pattern RAG}, for solving
the causality mining task using LLMs. Both the approaches leverage a fewshot
example repository---an external resource which we build (offline) for
supporting our proposed RAG approaches. The fewshot example repository contains
a collection of causal sentences with the cause and the effect phrases correctly
tagged. Each of these sentences in this repository are also indexed by a causal
connective. During causal inference using an LLM, given an input sentence both
the RAG approaches judiciously retrieve relevant examples dynamically from the
fewshot example repository, and augment the causality detection LLM prompts with
these examples for improving LLM performance. We conduct extensive experiments
on three datasets and five different LLMs to validate the performance of our
proposed RAG approaches against traditional supervised ML approaches and other
LLM prompting approaches, namely Zeroshot, Random fewshot and $k$NN RAG.
Experimental results show that our approaches outperform the baseline methods
for both causality detection and causality extraction tasks.



\begin{table}[h]
 \small   
\centering
\addtolength{\tabcolsep}{-0.01em}
\renewcommand{\arraystretch}{1.4} 
\begin{tabular}{@{} p{5.5cm}  p{1.9cm} @{} }
   \toprule
     Input Sentence &  \hspace{0.5em} Causal \newline Connectives \\ \toprule
   
\rowcolor{gray!10}However, as illustrated by these and other cases reported to date, the onset of <cause> troglitazone </cause> -induced <effect> liver injury </effect> is insidious and temporally variable. & \hspace{1.em} \textit{induced} 
\\

  When a <effect> tsunami </effect> is generated by a strong offshore <cause> earthquake </cause>, its first waves would reach the outer 
  coast minutes after the ground stops shaking. & \hspace{0.1em} \textit{is generated by}\\

 \rowcolor{gray!10} Highly viscous <cause> lavas </cause> lead to a violent <effect> eruption </effect>. & \hspace{1em}  \textit{lead to}\\
   
 \bottomrule

\end{tabular}
\caption{Causal sentences from the fewshot example repository with LLM generated causal connectives.}
\label{tab:few shot repository examples}
\vspace{-0.1in}
\end{table}


\begin{figure}[t]
 \small   

    \centering
    \includegraphics[width=2.5in]{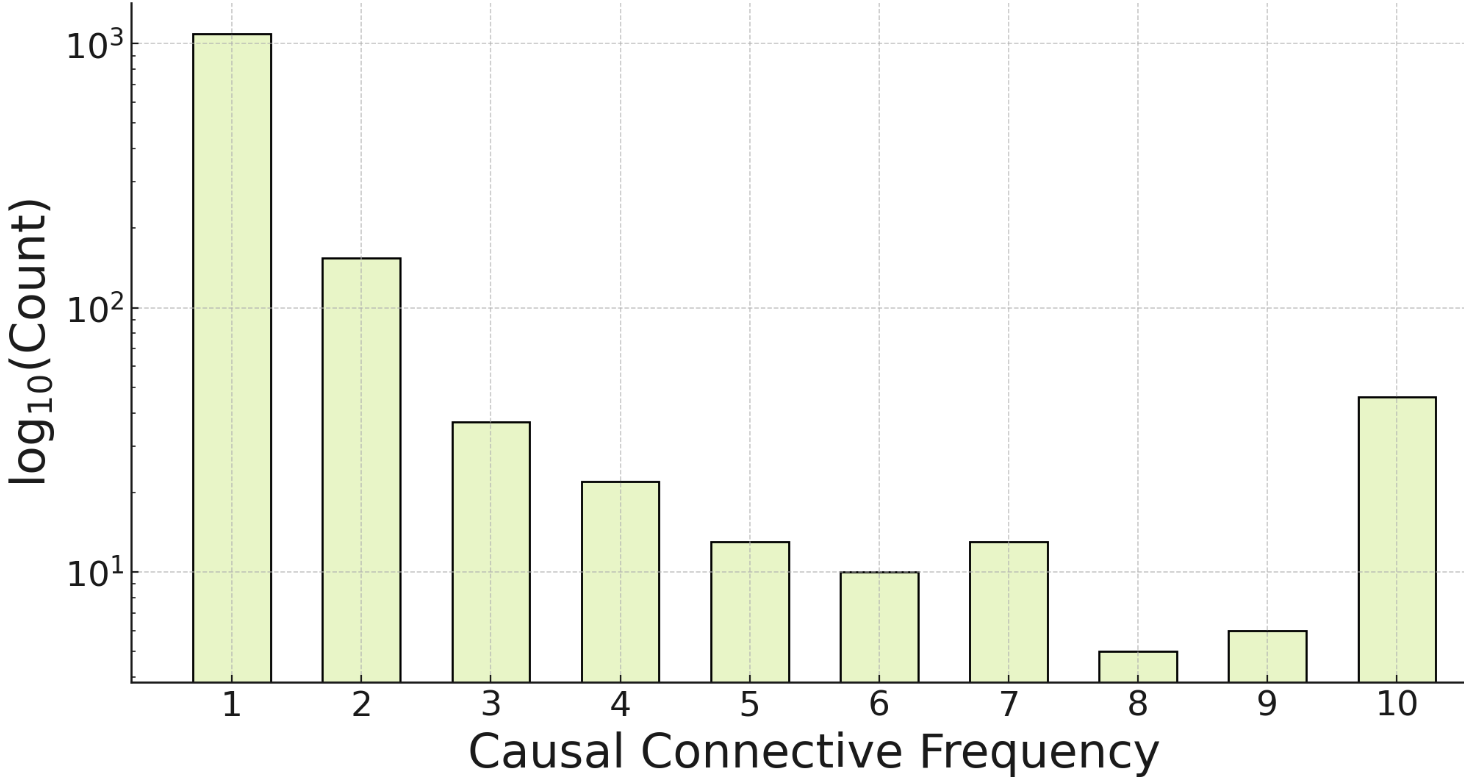}
    \caption{Logarithmic of the number of causal connectives in each frequency category ranging from 1 to 10.}
    \label{fig:hist_Causal_connective}
    \vspace{-0.23in}
\end{figure}

\begin{table*}
\small    
\centering
\setlength{\abovecaptionskip}{0.2em} 
\addtolength{\tabcolsep}{-0.01em}
\begin{tabular}{@{} p{.5cm} |p{15cm} @{}}

   \rowcolor{coc green} Freq. & Five Causal Connective sample\\
 \midrule
   \rowcolor{pastelgreen}1& {\fontfamily{cmtt}\selectfont generates, probable cause of, that caused the, as the consequence of, after the treatment by}\\

   \rowcolor{coc green} 2&{\fontfamily{cmtt}\selectfont led to, can lead to, may be induced by, effect, root causes of}\\

   \rowcolor{pastelgreen} 3& {\fontfamily{cmtt}\selectfont were caused by, derived from, experienced, was generated from, after initiation of, adverse effect of }\\

   \rowcolor{coc green} 4& {\fontfamily{cmtt}\selectfont caused the, after the use of, association of, causing, induces }\\

   \rowcolor{pastelgreen} 5& {\fontfamily{cmtt}\selectfont can induce, that resulted in, leading causes of, leading to, was the cause of } \\

   \rowcolor{coc green} 6& {\fontfamily{cmtt}\selectfont leads to, because of, ensued from, produces, can cause } \\

   \rowcolor{pastelgreen} 7& {\fontfamily{cmtt}\selectfont most common cause of, association between, as a complication of, side effect of, instigated by }\\

   \rowcolor{coc green} 8& {\fontfamily{cmtt}\selectfont result in, created by, as a result of, radiated, emits }\\

   \rowcolor{pastelgreen} 9& {\fontfamily{cmtt}\selectfont resulting from, results in, had caused, radiating from, had caused }\\

  \rowcolor{coc green} 10 & {\fontfamily{cmtt}\selectfont triggered by, induced by, arose from, is one of the main causes of, -associated }\\


\end{tabular}
\caption{Causal connective examples for each frequency category}
\label{tab:Causal connective examples for each frequency category}
\end{table*}


\begin{figure*}[h]
 \small   
\setlength{\abovecaptionskip}{0.1em} 
    \centering
    \includegraphics[width=4.5in]{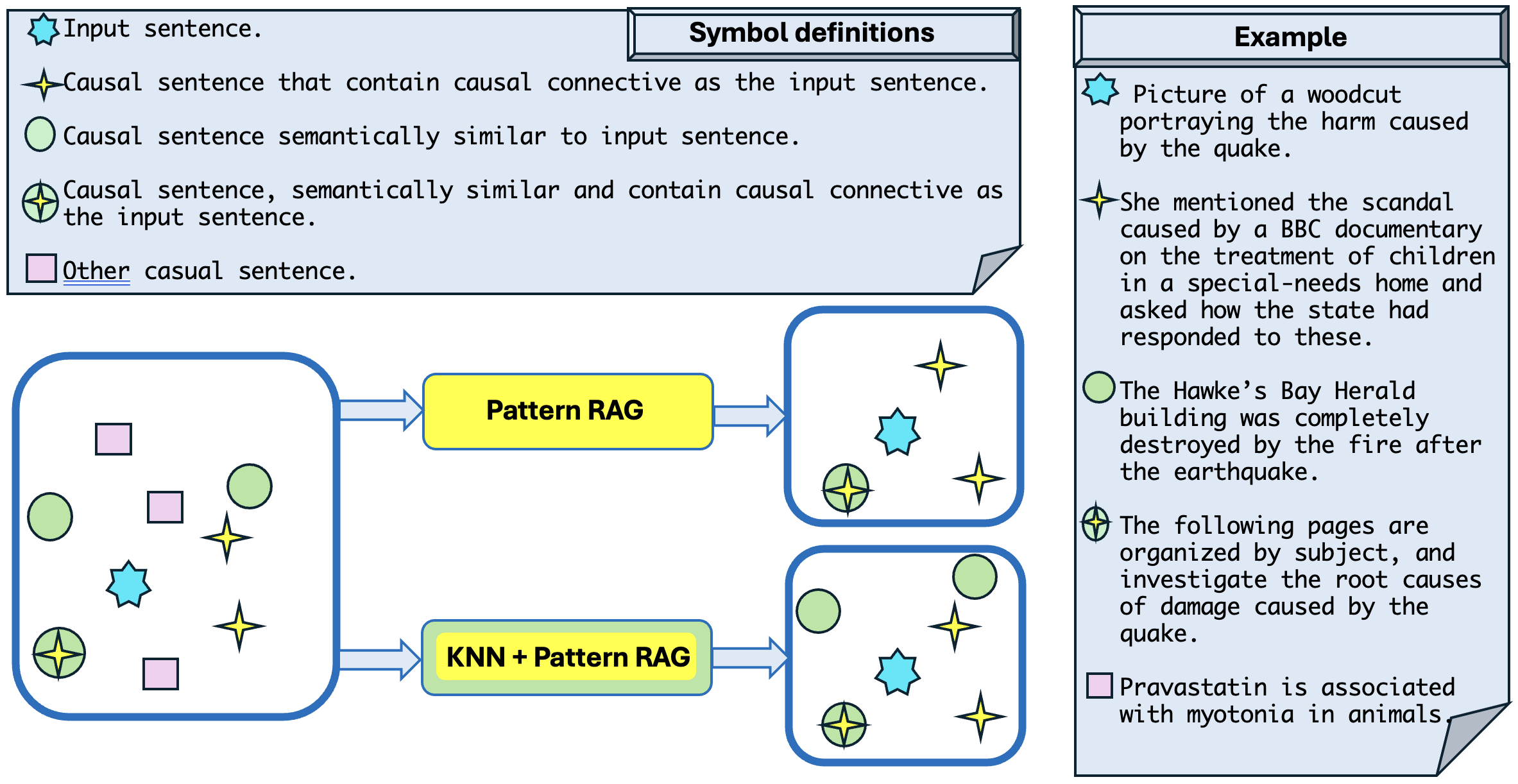}
    \caption{Dynamic fewshot selection mechanism of Pattern RAG and $k$NN+Pattern RAG. An example sentence from SemEval dataset and related sentences that are selected by different RAG schemes are also shown.}
    \vspace{-0.2in}
    \label{fig:Model}
\end{figure*}

\section{Related Works}

Causality mining research in the past few decades can be broadly categorized into three main approaches: rule-based (non-statistical), traditional machine learning-based, and deep learning-based.  
Rule-based approaches~\cite{kaplan1991knowledge, joskowicz1989deep, kontos1991acquisition, garcia1997coatis, girju2002text, khoo1998automatic, sadek2013automatic} mainly use different linguistic patterns  
and causality connectives (e.g., ``caused'', ``lead to'' and ``triggered'') to detect causality. These methods require significant human intervention for identifying the rules. Often, rules extracted from one domain may not work for another domain as effectively. 
Traditional machine learning approaches~\cite{girju2002text, bethard2008learning, pal2010ju, sorgente2013automatic, pakray2014open} are more advantageous than rule-based approaches as they require less human effort. These methods automate the pattern extraction process by utilizing  NLP tools like WordNet, Google N-grams, POS tagging, and apply them in models like decision trees, SVM, and Naive Bayes for causality extraction. 
 
The advent of deep learning models and the attention mechanism help overcome the feature sparsity problem and made the training more effective~\cite{de2017causal, ponti2017event, dasgupta2018automatic, wu2019enriching, zhao2019improving, ali2021causality, tian2022improving, lan2023modeling, chen2021zs}. Prior studies use various deep learning architectures, including Self-attentive BiLSTM-CRF with Transferred Embeddings (SCITE)~\cite{li2021causality}, linguistically informed Bi-LSTM model ~\cite{dasgupta2018automatic}, bidirectional LSTM with a CRF layer (BI-LSTM-CRF)~\cite{huang2015bidirectional} and dependency-aware transformer based model (DEPBERT)~\cite{kabir2023deep}. However, achieving better performance with deep learning models requires high computational resources and a large training corpus, preferably from the same domain, which poses a significant obstacle. 
The advance of LLMs and prompt engineering helps to overcome the limitations of deep learning models. \citet{zhang2024causal} utilize the idea of RAG-based LLMs to deduce causal relationships from a large corpus of scientific literature for the causal graph discovery task. 

\citet{liu2021makes} introduce the idea of selecting dynamic few-shot examples for a test instance based on similarity measures. They experiment these $k$NN based in-context example selection method for sentiment classification, table-to-text generation, and question answering tasks. 
Similar approaches are also used for relation extraction tasks \cite{nori2023can, liu2023freda, efeoglu2024retrieval}. We make minor modification to \citet{liu2021makes} to adapt it our task and use it as one of the baseline.

$k$NN-based method chooses semantically similar examples, but they are not always helpful for cause effect phrase extraction task. Specifically, when the phrases are not identified by the semantic meaning of entire sentence, but by causal connectives, such a method performs poorly.
\citet{wang2023learning} utilizes a retriever trained on a labeled dataset, which involves a costly training phase. In contrast, our pattern-based method does not require such a retriever, offering a more efficient and lightweight alternative.
\citet{zhang2022active} uses reinforcement learning to iteratively train a policy, which demands significant time and computational resources.
\citet{li2023unified} proposes a Unified Demonstration Retriever (UDR) that retrieves examples across diverse tasks using a multi-task listwise ranking framework, reducing storage and deployment costs compared to task-specific retrievers.
These works have explored the in context example selection methods but haven't explored the causality mining tasks. Since cause-effect phrase extraction is a sequence labeling task, incorporating these baselines requires to adapt these methods substantially, which in-itself can be a new research direction. 

\section{Methodology}
Inspired by recent advancements in Retrieval-Augmented Generation (RAG) demonstrating the efficacy of dynamically selected few-shot examples for enhancing LLM performance (Liu et al., 2021), we propose a novel approach to causality mining that utilizes an external knowledge base to augment LLM prompts. We construct a concise knowledge base of causal examples, and develop two RAG-based methods---\textbf{Pattern RAG} and \textbf{$k$NN+Pattern RAG}---to dynamically select relevant instances for a given task. These approaches retrieve in-context examples from our fewshot example repository by leveraging causal connectives and sentence embedding similarity.

\subsection{\textbf{Fewshot Example Repository Creation}}
The fewshot example repository is created off-line, which is to be used as an
external source to support retrieval-augmented prompt generation with a large
language model. The desiderata of this repository, which stores example
causality sentences are as follows: (1) it should contain examples similar to
any test instance provided by a user; (2) the examples should be domain-neutral;
and (3) they should be concise.

To fulfill these requirements, we first combine all the causal sentences from the training datasets of SemEval, ADE and \citeauthor{li2021causality} (4,082 sentences in total). Each sentence is then fed into GPT-3.5-turbo with an ICL prompt to identify 
the causal connectives. 
Some sample sentences with their identified causal connectives (from GPT-3.5-turbo) are shown in Table \ref{tab:few shot repository examples}.

\begin{table*}[h]
\centering
\small
\renewcommand{\arraystretch}{1.4}

\addtolength{\tabcolsep}{-0.3em}
\begin{tabular}{p{1.1cm}| p{2.5cm} || p{0.75cm} p{0.75cm} p{0.75cm} p{0.75cm} | p{0.75cm} p{0.75cm} p{0.75cm} p{0.75cm} | p{0.75cm} p{0.75cm} p{0.75cm} p{0.75cm}}

   \hline
     LLM  & Method & \multicolumn{4}{c|}{SemEval} & \multicolumn{4}{c|}{ADE} & \multicolumn{4}{c}{\citeauthor{li2021causality} dataset} \\
    \cline{3-14}
          &        &Acc. & F1 & P & R &Acc.& F1 & P & R& Acc.& F1 & P & R \\
    \hline
    \hline
   GPT-3.5& Zeroshot & 0.68 & 0.73 &0.60 &\textbf{0.93}&   0.76	&0.81	&0.72	&\textbf{0.93}& 0.65	&0.57	&0.40	&0.95\\
    -turbo&Random fewshot&0.85 & \textbf{0.84} & 0.83 & 0.85&0.73	&0.67	&\textbf{0.97}	&0.51 &  0.89	&0.72	&\textbf{0.86}	&0.63\\
    &$k$NN RAG & 0.84& 0.81& 0.88&0.75 &0.85&0.84&0.92&0.78  &0.91 &0.84 &0.77 &0.93\\
      &  \cellcolor{gray!10}Pattern RAG      & \cellcolor{gray!10} \textbf{0.86}&	\cellcolor{gray!10}0.83&	\cellcolor{gray!10}\textbf{0.90}&\cellcolor{gray!10}0.78 & \cellcolor{gray!10}0.86	&\cellcolor{gray!10}0.85	&\cellcolor{gray!10}0.96	&\cellcolor{gray!10}0.77 &  \cellcolor{gray!10}\textbf{0.93}	&\cellcolor{gray!10}\textbf{0.86}	&\cellcolor{gray!10}0.85	&\cellcolor{gray!10}0.87 \\
     &\cellcolor{gray!10} $k$NN+Pattern RAG& \cellcolor{gray!10}\textbf{0.86}&\cellcolor{gray!10} \cellcolor{gray!10}\textbf{0.84} &\cellcolor{gray!10}0.89&\cellcolor{gray!10}0.80& \cellcolor{gray!10}\textbf{0.88}& \cellcolor{gray!10}\textbf{0.88}& \cellcolor{gray!10}0.92&\cellcolor{gray!10}0.85 &\cellcolor{gray!10} 0.92&\cellcolor{gray!10} 0.85& \cellcolor{gray!10}0.76 &\cellcolor{gray!10}\textbf{0.96}\\

 \hline
 GPT-4o& Zeroshot& 0.85 & 0.83 &0.83 &0.84  &   0.81	&0.81	&0.89	&0.75  & 0.89	&0.79	&0.72	&0.88\\
  &Random fewshot& 0.87 & 0.84 & 0.90 & 0.79 &0.85	&0.84	&0.95	&0.76 &  0.91	&0.82	&0.86	&0.78 \\
    &$k$NN RAG & 0.90 & 0.88 & \textbf{0.92} & 0.84 & 0.85 & 0.84&  0.95& 0.76 & \textbf{0.96} & 0.92 & 0.87 &0.97\\

    & \cellcolor{gray!10} Pattern RAG     & \cellcolor{gray!10} 0.88&\cellcolor{gray!10}	0.86&\cellcolor{gray!10}	0.89&\cellcolor{gray!10}0.84 &\cellcolor{gray!10} 0.89
    &\cellcolor{gray!10}0.89	&\cellcolor{gray!10}0.95	&\cellcolor{gray!10}\textbf{0.84} &  \cellcolor{gray!10}0.94	&\cellcolor{gray!10}0.88	&\cellcolor{gray!10}0.85	&\cellcolor{gray!10}0.92\\
    & \cellcolor{gray!10}$k$NN+Pattern RAG & \cellcolor{gray!10}\textbf{0.91} &\cellcolor{gray!10}\textbf{0.90}&\cellcolor{gray!10}\textbf{0.92}& \cellcolor{gray!10}\textbf{0.87} &\cellcolor{gray!10}\textbf{0.90} &\cellcolor{gray!10} \textbf{0.90}&\cellcolor{gray!10}\textbf{0.96}&\cellcolor{gray!10}\textbf{0.84}&\cellcolor{gray!10}\textbf{0.96}& \cellcolor{gray!10}\textbf{0.93}&\cellcolor{gray!10}\textbf{ 0.89}&\cellcolor{gray!10}\textbf{0.98} \\

 \hline
 Llama3& Zeroshot & 0.79 & 0.79 &0.70 &0.91 &   0.83	&0.85	&0.78	&0.93 & 0.73	&0.62	&0.47	&0.94\\
  -8b&Random fewshot& 0.78 & 0.74 & 0.78 & 0.71 &0.87	&0.88	&0.87	&0.90 & \textbf{0.84}	&0.72	&\textbf{0.63}	&0.83\\
    &$k$NN RAG & 0.72&0.75&0.62&\textbf{0.96}& 0.74 & 0.80&0.69 & 0.95&0.57&0.53&0.36&0.98\\

& \cellcolor{gray!10} Pattern RAG     &\cellcolor{gray!10}\textbf{0.83}&\cellcolor{gray!10}	\textbf{0.83}&\cellcolor{gray!10}	\textbf{0.79}&\cellcolor{gray!10}0.86 &\cellcolor{gray!10} \textbf{0.88}	&\cellcolor{gray!10}\textbf{0.89}	&\cellcolor{gray!10}\textbf{0.85}	&\cellcolor{gray!10}\textbf{0.93} &\cellcolor{gray!10}  0.83	&\cellcolor{gray!10}\textbf{0.73}	&\cellcolor{gray!10}0.59	&\cellcolor{gray!10}0.94 \\
 & \cellcolor{gray!10}$k$NN+Pattern RAG&\cellcolor{gray!10} 0.82 & \cellcolor{gray!10}0.82 &\cellcolor{gray!10}0.77 &\cellcolor{gray!10}0.87 &\cellcolor{gray!10}0.81 &\cellcolor{gray!10}0.83 &\cellcolor{gray!10} 0.80&\cellcolor{gray!10}0.86 &\cellcolor{gray!10}0.74 &\cellcolor{gray!10}0.64 &\cellcolor{gray!10} 0.48&\cellcolor{gray!10}\textbf{ 0.95} \\

 \hline
 Gemma2& Zeroshot& 0.81 & 0.77 &0.82 &0.73  &   0.67	&0.59	&0.88	&0.45 & 0.89	&0.77	&0.79	&0.75 \\
 -9b-it &Random fewshot& 0.79 & 0.75 & 0.84 & 0.67  &\textbf{0.71}	&\textbf{0.66}	&\textbf{0.91}	&\textbf{0.52} & 0.88	&0.73	&0.79	&0.68\\
    &$k$NN RAG & 0.83&0.80&\textbf{0.90}&0.71 &0.60&0.43&0.86&0.28&\textbf{0.92}&\textbf{0.85}& 0.85&0.85\\
    &\cellcolor{gray!10}  Pattern RAG      &\cellcolor{gray!10} 0.80&\cellcolor{gray!10}	0.76&\cellcolor{gray!10}	0.85&\cellcolor{gray!10}0.69&\cellcolor{gray!10} 0.64	&\cellcolor{gray!10}0.51	&\cellcolor{gray!10}0.89	&\cellcolor{gray!10}0.36 &\cellcolor{gray!10}  0.91	&\cellcolor{gray!10}0.80	&\cellcolor{gray!10}\textbf{0.85}	&\cellcolor{gray!10}0.76\\
      & \cellcolor{gray!10}$k$NN+Pattern RAG& \cellcolor{gray!10}\textbf{0.83} &\cellcolor{gray!10}\textbf{ 0.81} &\cellcolor{gray!10} 0.84 &\cellcolor{gray!10}\textbf{0.78}&\cellcolor{gray!10}0.65 & \cellcolor{gray!10}0.56& \cellcolor{gray!10}0.85& \cellcolor{gray!10}0.42&\cellcolor{gray!10}0.91  & \cellcolor{gray!10}0.84&\cellcolor{gray!10}0.78& \cellcolor{gray!10}\textbf{0.90}\\

 \hline

  Mixtral-& Zeroshot& 0.79& 0.78& 0.72& 0.86 &0.78 &0.80 &0.78 &0.82 & 0.76 &0.64 & 0.50 & 0.88\\
 8x7b &Random fewshot& 0.75 &0.69& 0.77 & 0.63 &0.72 &0.70 &\textbf{ 0.84}& 0.60& 0.81 &0.64 &\textbf{0.62} & 0.65\\
    &$k$NN RAG & 0.81& 0.80&\textbf{ 0.79}& 0.81&0.78&0.78&\textbf{0.84}&0.73&\textbf{0.84}&\textbf{0.75}&0.61&0.96\\
    &\cellcolor{gray!10}  Pattern RAG     &\cellcolor{gray!10}\textbf{0.83}&\cellcolor{gray!10}\textbf{0.81}&\cellcolor{gray!10} 0.78 &\cellcolor{gray!10}0.85 &\cellcolor{gray!10}0.80 &\cellcolor{gray!10} 0.81 &\cellcolor{gray!10} 0.81&\cellcolor{gray!10}0.81&\cellcolor{gray!10} 0.82&\cellcolor{gray!10} 0.72&\cellcolor{gray!10}0.58&\cellcolor{gray!10}0.93\\
    &\cellcolor{gray!10}$k$NN+Pattern RAG&\cellcolor{gray!10}0.80&\cellcolor{gray!10}0.80&\cellcolor{gray!10}0.71&\cellcolor{gray!10}\textbf{0.92} &\cellcolor{gray!10}\textbf{0.81} &\cellcolor{gray!10} \textbf{0.84}&\cellcolor{gray!10} 0.77&\cellcolor{gray!10}\textbf{ 0.92}&\cellcolor{gray!10} 0.75 & \cellcolor{gray!10}0.66 &\cellcolor{gray!10} 0.49 &\cellcolor{gray!10} \textbf{0.99}\\

 \hline

\end{tabular}
\caption{Causality detection results.}
\label{tab:Causality detection results}
\vspace{-0.2in}
\end{table*}


The sentences are then indexed by the causal connectives identified by GPT, and stored in a data structure, called \textit{Fewshot Example DB}. Since our experiment does not require many examples,  and also to make the repository memory efficient, we keep up to 10 random examples per unique causal connective. Note that some example sentences are shared by multiple causal connectives, as GPT does not always extract the
exact casual connectives; for instance, it may extract ``caused by'' in one instance and ``caused by the'' in another, resulting in two different connectives. At the end, our \textit{Fewshot Example DB} contains 2,365 instances indexed by 1,394 unique causal connectives, with 80  connectives having at least 5 examples, while the rest occur 1 to 5 times. Figure \ref{fig:hist_Causal connective} shows the causal connectives count in each frequency category of \textit{Fewshot example DB}. 
We manually verify the quality of causal connectives by randomly selecting five causal connective from each frequency category, ranging from 1 to 10, in our repository. This process is repeated over several iterations, and one of the results is shown in Table \ref{tab:Causal connective examples for each frequency category}. 


\subsection{\textbf{{RAG Prompts for Causality Mining}}}

LLMs offer powerful language understanding and generation capabilities, but they are not reliable information sources and they lack access to information beyond their training data. Retrieval-augmented generation enables LLMs to witness in-context examples
which are highly relevant to the given task. These in-context examples are chosen
by RAG approaches. In this work,
for the causality extraction task, we propose two RAG ideas: Pattern RAG and $k$NN+Pattern RAG. 

\noindent \textbf{Pattern RAG:} 
This is our first RAG scheme, where the fewshot examples are chosen by matching the causal connective of the input sentence with those of the sentences in the textit{Fewshot Example DB}. For example, for input sentence ``fever is caused by flu'', fewshot examples in the repository which has ``cause by'' connective are chosen for prompt augmentation. 
Since causal connective detection is not always exact, we select those sentences for which the causal connective is more than 90 percent similar to the causal connective of the input. 
If we find more than 10 matched examples, we randomly choose 10 of them and filter out the rest. This Pattern RAG scheme gets similar fewshot examples based on the presence of similar causal connectives in the examples. Our hypothesis is that, with these fewshot examples, LLM will be able to align the causal connective between the input sentence and the fewshot example, and be able to identify the cause and effect phrases more effectively. In Figure~\ref{fig:Model}, the input sentence has the ``caused by'' pattern, the retrieved fewshot examples also have ``caused by'' 
in them (see the sentence tagged with a yellow star).

\noindent \textbf{$k$NN+Pattern RAG:}
In this RAG scheme we first identify causal sentences that are semantically similar to the input. We use “text-embedding-ada- 002” model to obtain vector representations of the input sentence (test instances) and all example instances in our Fewshot Example DB. The k-nearest neighbor search algorithm is applied to find the 10 most similar examples to the given test instance from the DB. For
$k$NN+Pattern RAG, we concatenate the 10 examples identified by
$k$NN RAG and those retrieved by Pattern RAG, resulting in a total of 20 examples for fewshot prompt augmentation.
$k$NN RAG retrieves 10 examples that are semantically
similar to the input sentence, while Pattern RAG retrieves all the examples that
show a causality relation with the same causal connective present in the input
sentence. Our hypothesis is that this combination will help the LLM by
leveraging both  sentential semantics and causal pattern syntax.
Figure~\ref{fig:Model} provides a  pictorial depiction of both our RAG approaches.




\begin{table*}[h]
\centering
\small 
\setlength{\abovecaptionskip}{0.1em} 
 
\renewcommand{\arraystretch}{1.4}
\addtolength{\tabcolsep}{-0.3em}
\begin{tabular}{p{1.5cm}| p{2.5cm} || p{1.3cm}  |  p{1.3cm}  |  p{1.1cm} p{1.2cm} p{1.1cm}}
     \hline

     LLM  & Method & SemEval & ADE & \multicolumn{3}{c}{\citeauthor{li2021causality} dataset} \\
    \cline{3-7}
          &         & Accuracy  & Accuracy &  F1 & Precision & Rrecall \\
    \hline
    \hline
    DEPBERT& Supervised& 0.65 & 0.61& 0.12& 0.12&0.13\\

    \hline

   GPT-3.5& Zeroshot & 0.85	&  0.75	 &  \textbf{0.54}	&0.57	&\textbf{0.52}\\
    -turbo&Random fewshot&  0.83	 &0.79	 & 0.47	&0.56	&0.40\\
    &$k$NN RAG &  \textbf{0.91}& \textbf{0.83}& 0.46 & 0.60 &0.38\\

    & \cellcolor{gray!10} Pattern RAG     &\cellcolor{gray!10} 0.90	 &\cellcolor{gray!10}0.80	&\cellcolor{gray!10} 0.52	&\cellcolor{gray!10}0.61	&\cellcolor{gray!10}0.46\\
    &\cellcolor{gray!10} $k$NN+Pattern RAG&\cellcolor{gray!10}\textbf{0.91}&\cellcolor{gray!10}0.82&\cellcolor{gray!10}0.51&\cellcolor{gray!10}\textbf{0.62}&\cellcolor{gray!10}0.44 \\

 \hline
 GPT-4o& Zeroshot& 0.82	 &  0.66 &  0.76	&0.79	&0.73 \\
  &Random fewshot& 0.82	  &0.77	  &0.75	&0.77	&0.73\\
    &$k$NN RAG & 0.87  &0.79& 0.77&0.83&0.72 \\
    & \cellcolor{gray!10} Pattern RAG     &\cellcolor{gray!10} 0.89	 &\cellcolor{gray!10}0.83 &\cellcolor{gray!10} \textbf{0.80}	&\cellcolor{gray!10}\textbf{0.83}	&\cellcolor{gray!10}\textbf{0.78}\\
    &\cellcolor{gray!10} $k$NN+Pattern RAG&\cellcolor{gray!10}\textbf{0.89}&\cellcolor{gray!10}\textbf{0.84}&\cellcolor{gray!10}0.75&\cellcolor{gray!10}0.78&\cellcolor{gray!10}0.73 \\

 \hline
 Llama3-8b& Zeroshot & 0.77	 &  0.75	 &  0.65	&0.84	&0.53 \\
  &Random fewshot& 0.76	&\textbf{0.78}  &0.65	&\textbf{0.85}	&0.52\\
    &$k$NN RAG & \textbf{0.83}&0.76  & 0.64 & 0.82 & 0.52\\
    & \cellcolor{gray!10} Pattern RAG     & \cellcolor{gray!10}\textbf{0.83}	&\cellcolor{gray!10}0.77	  &\cellcolor{gray!10} 0.64	&\cellcolor{gray!10}0.83	&\cellcolor{gray!10}0.53\\
    &\cellcolor{gray!10} $k$NN+Pattern RAG&\cellcolor{gray!10}\textbf{0.83}&\cellcolor{gray!10}0.75&\cellcolor{gray!10}\textbf{0.66}&\cellcolor{gray!10}0.83&\cellcolor{gray!10}\textbf{0.55} \\

 \hline
 Gemma2& Zeroshot & 0.76 &0.76	&0.64	&0.76	&0.55\\
 -9b-it &Random fewshot& 0.77 &0.81	&0.68	&0.83	&0.58\\
    &$k$NN RAG & 0.83&0.83  & 0.64 &0.81 & 0.53\\
    & \cellcolor{gray!10} Pattern RAG     & \cellcolor{gray!10}0.83 &\cellcolor{gray!10}0.83	&\cellcolor{gray!10}0.72	&\cellcolor{gray!10}0.82	&\cellcolor{gray!10}0.61\\
    &\cellcolor{gray!10} $k$NN+Pattern RAG&\cellcolor{gray!10}\textbf{0.87}&\cellcolor{gray!10}\textbf{0.84}&\cellcolor{gray!10}\textbf{0.74}&\cellcolor{gray!10}\textbf{0.90}&\cellcolor{gray!10}\textbf{0.63} \\

 \hline
 
  Mixtral-& Zeroshot&0.73&0.71&0.66&0.81&\textbf{0.56}  \\
 8x7b &Random fewshot& 0.74&0.79&0.66& 0.84&0.55\\
    &$k$NN RAG &\textbf{0.81}&0.80&0.62&0.80&0.51 \\
    & \cellcolor{gray!10} Pattern RAG     &\cellcolor{gray!10}\textbf{0.81}  &\cellcolor{gray!10}\textbf{0.83}&\cellcolor{gray!10}\textbf{0.67}&\cellcolor{gray!10}\textbf{0.86}&\cellcolor{gray!10}0.55\\
    &\cellcolor{gray!10} $k$NN+Pattern RAG&\cellcolor{gray!10}0.80&\cellcolor{gray!10}0.81&\cellcolor{gray!10}0.66&\cellcolor{gray!10}0.80&\cellcolor{gray!10}\textbf{0.56} \\

 \hline

\end{tabular}
\caption{Causality extraction results.}
\label{tab:Cause-Effect_phrase_extraction_results}
\vspace{-0.2cm}
\end{table*}


\section{Experiments}
We test our proposed RAG approaches with five different LLMs on three datasets, comparing our results for two tasks against competing methods. The first task is causality detection, a binary classification task. The input to this task consists of sentences labeled as 1 (causal sentence) and 0 (non-causal sentence). The LLM outputs a response of 1 if the sentence has a causality relation, otherwise it outputs 0. For evaluation, 
We use standard classification evaluation metrics: accuracy, F1 score, precision and recall. 

The second task is causality extraction, a sequence labeling task where the LLM identifies cause and effect phrases from the input sentence. The labeled dataset in our experiments typically marks single words as cause and effect. However, in real-world scenarios, more detailed phrase representations of cause and effect are often more informative than single words. To accommodate this, our designed prompts guide the LLM to extract phrases instead of single words when identifying cause and effect within a sentence. For example, in the input sentence, ``\textit{The truck carried homemade weapons, and the blast was caused by the mishandling of weapons},'' the ground truth dataset labels ``\textit{mishandling}'' as the cause and ``\textit{blast}'' as the effect. However, with our prompts, the LLM generates ``\textit{mishandling of weapons}'' 
as the cause and ``\textit{blast}'' as the effect. Additionally, we explicitly instruct the LLM in the prompt to ensure there is no overlap  between the predicted cause and effect phrases.

To evaluate the performance of causality extraction task, we use different metrics
based on whether a sentence has only one cause-effect phrase or multiple cause-effect phrases in the ground truth data. For the SemEval and ADE datasets, each sentence has only a single set of cause-effect phrase; so for these datasets we only check whether
the predicted cause and effect phrases match with the ground truth
cause and effect phrases, respectively. If the ground truth word is presented in the predicted phrases (for both cause and effect), we consider the causality extraction task on that sentence a success. Accuracy is simply the fraction of sentences
for which causality extraction is a success. Note that, for these two datasets, we 
add a constraint in the prompt so that LLM extracts only one cause-effect pair from each sentence. 

The \citeauthor{li2021causality} dataset have sentences which have multiple cause-effect phrases, so for this dataset accuracy is hard to produce. So we use precision, recall, and F1 metrics as below. From all the sentences in \citeauthor{li2021causality},  we first create triplets $(S, C, E)$, each denoting a distinct cause ($C$) and effect ($E$) phrase pair in the sentence $S$. Let the set of these triplets be called the triplet test set ($T_t$). Say, $T_p$ is the collection of triplets formed from the LLM for all the sentences. Then, precision of the LLM model for the \citeauthor{li2021causality} dataset can be calculated as $P = |T_t \cap T_p|/ |T_p|$, and recall can be calculated as $R = |T_t \cap T_p|/ |T_t|$. From $P$ and $R$, we then
compute the $F1$ value. While computing intersection between $T_p$ and $T_t$, 
a predicted triplet in $T_p$ is considered a match with a ground truth triplet in $T_t$,
when their sentences are identical, and the ground truth cause and effect phrases are present in the predicted cause and effect phrase.

\subsection{{\bf Baseline Methods}}
As baseline methods, we use two static prompt strategies: Zeroshot and Random fewshot for both the detection and extraction tasks. And a dynamic in context example selection baseline, $k$NN RAG. 
For the causality extraction task, we use the three deep learning based methods Bi-LSTM (Dasgupta), Bi-LSTM-CRF (SCITE) and DEPBERT \citep{kabir2023deep} which does not use LLM.


\noindent 
\textbf{Zeroshot:}
In this baseline approach for both causality detection  and extraction, we provide an ICL prompt (shown in Appendix) that combines the concepts of causal connectives with a detailed task explanation for the model. Experiments conducted under Zeroshot setting on different LLMs demonstrate the performance of LLMs on causality mining tasks when no prior examples are available.

\noindent 
\textbf{Random fewshot: }
For this baseline approach, we randomly select 10 examples from \textit{Fewshot Example DB} and concatenate them with the Zeroshot setting prompt for both tasks. For causality detection, an input example includes only the sentence with cause and effect phrases whereas for causality extraction,  we additionally tag the causal connective of the example. Note that, the prompts and example formats for Random fewshot, $k$NN RAG, Pattern RAG, and $k$NN+Pattern RAG are the same, differing only in the choice of example selections.

\noindent
\textbf{$k$NN-RAG:}
In existing RAG works [35], examples that are semantically similar to the input are chosen to augment the LLM prompt. We adapt this approach as a baseline for causality mining and call it as $k$NN RAG.
We apply the k-nearest neighbor search algorithm on the vector representations to find the 10 most similar examples to the given test instance from the DB. We then concatenate these examples with our prompts for causality detection and extraction. 
In Figure \ref{fig:Model}, we see that for an input sentence related to destruction caused by an earthquake, the $k$NN RAG selects the examples with similar meanings (shown in green circles).

\noindent 
\textbf{Dasgupta (Bi-LSTM): } \citet{dasgupta2018automatic} proposed one of the earliest deep neural network-based method for causality extraction. Their approach combines embeddings from Word2Vec with a linguistic feature vector, which are then input to a bi-directional LSTM (bi-LSTM) model.

\noindent 
\textbf{SCITE (Bi-LSTM-CRF): } \citet{li2021causality} proposed a causality extraction model that leverages a BiLSTM CRF back bone, enhanced with Flair embeddings and multihead self-attention.

\noindent 
\textbf{DEPBERT (Transformer): }\citep{kabir2023deep}
DEPBERT is a state-of-the-art transformer-based supervised model that leverages the dependency tree of a sentence on top of a BERT model to extract cause and effect phrases.


\subsection{Datasets}
For our experiments, we use three well-known causality datasets: SemEval \cite{hendrickx2019semeval}, ADE \citep{gurulingappa2012development}, and  \citeauthor{li2021causality} dataset\cite{li2021causality} . Their detailed discussion is
provided in Appendix. Statistics of the dataset is provided in Table~\ref{tab:Dataset Statistics} (in Appendix). Table \ref{tab:Task1 Causality detection input examples} and \ref{tab:Task2: Cause-Effect phrase extraction input examples} shows few example inputs for causality detection and extraction tasks, respectively.

\begin{table}[h]
 \small   
\centering
\addtolength{\tabcolsep}{-0.01em}
\begin{tabular}{p{4cm}  p{3cm}  }
\toprule

    Methods & Accuracy \\
    \midrule
   
 \rowcolor{gray!10} Dasgupta (Bi-LSTM) & 0.78\\

  SCITE (Bi-LSTM-CRF) & 0.78\\

 \rowcolor{gray!10} DEPBERT (Transformer) & 0.86\\
   
 \textbf{Pattern RAG } & \textbf{0.88}\\
   \rowcolor{gray!10} \textbf{$k$NN+Pattern RAG} & \textbf{0.90}\\
\bottomrule


\end{tabular}
\caption{Accuracy comparison of our RAG methods (GPT-3.5-turbo Pattern RAG and $k$NN+Pattern RAG) with deep learning methods on SemEval dataset.}
\label{tab:DL methods comp}
\vspace{-0.2in}
\end{table}

\section{Results}

Table \ref{tab:DL methods comp} shows the comparison of three deep learning-based baseline methods trained on the entire SemEval train dataset with our RAG based methods, evaluated on the SemEval test set as used in the DEPBERT paper. Additionally, the best-performing deep learning method, DEPBERT, has been included in Table \ref{tab:Cause-Effect_phrase_extraction_results} to compare its performace against all the LLM results. This DEPBERT result is obtained by training on our few-shot DB and testing on the cleaned test datasets, as detailed in Table \ref{tab:Dataset Statistics}.
Our experiment results for the causality detection and extraction tasks are shown in Table~\ref{tab:Causality detection results} and Table~\ref{tab:Cause-Effect_phrase_extraction_results}, respectively.
From these tables, we can see that our RAG-based methods demonstrate significant performance improvements over the baseline methods across all three datasets.

For causality detection, the $k$NN+Pattern RAG method achieves the best performance when paired with the GPT-4o model. The corresponding best baseline method for GPT-4o across all three datasets is $k$NN RAG. The F1 improvements by the best of our RAG approach over the best baseline approach for SemEval, ADE, and \citeauthor{li2021causality} are +2\%, +7\%, and +1\%, respectively. 

For causality extraction, the SemEval dataset shows its best performance with the GPT-3.5-turbo model in the $k$NN RAG and $k$NN+Pattern RAG settings. The ADE and \citeauthor{li2021causality} datasets achieve their best performance with the GPT-4o model in $k$NN+Pattern RAG and Pattern RAG, respectively. 
Our best-performing RAG model shows no improvement in accuracy over the best baseline ($k$NN RAG) for the SemEval dataset, maintaining an accuracy of 0.91. However, significant improvements are observed for the ADE and \citeauthor{li2021causality} datasets. For ADE, the GPT-4o model using $k$NN+Pattern RAG improves accuracy by +6\% (0.84 vs. 0.79). In the \citeauthor{li2021causality} dataset, the Pattern RAG method boosts F1 by +4\% (0.80 vs. 0.77).

\begin{figure}[t]
 \small   

    \centering
    \includegraphics[width=2.5in]{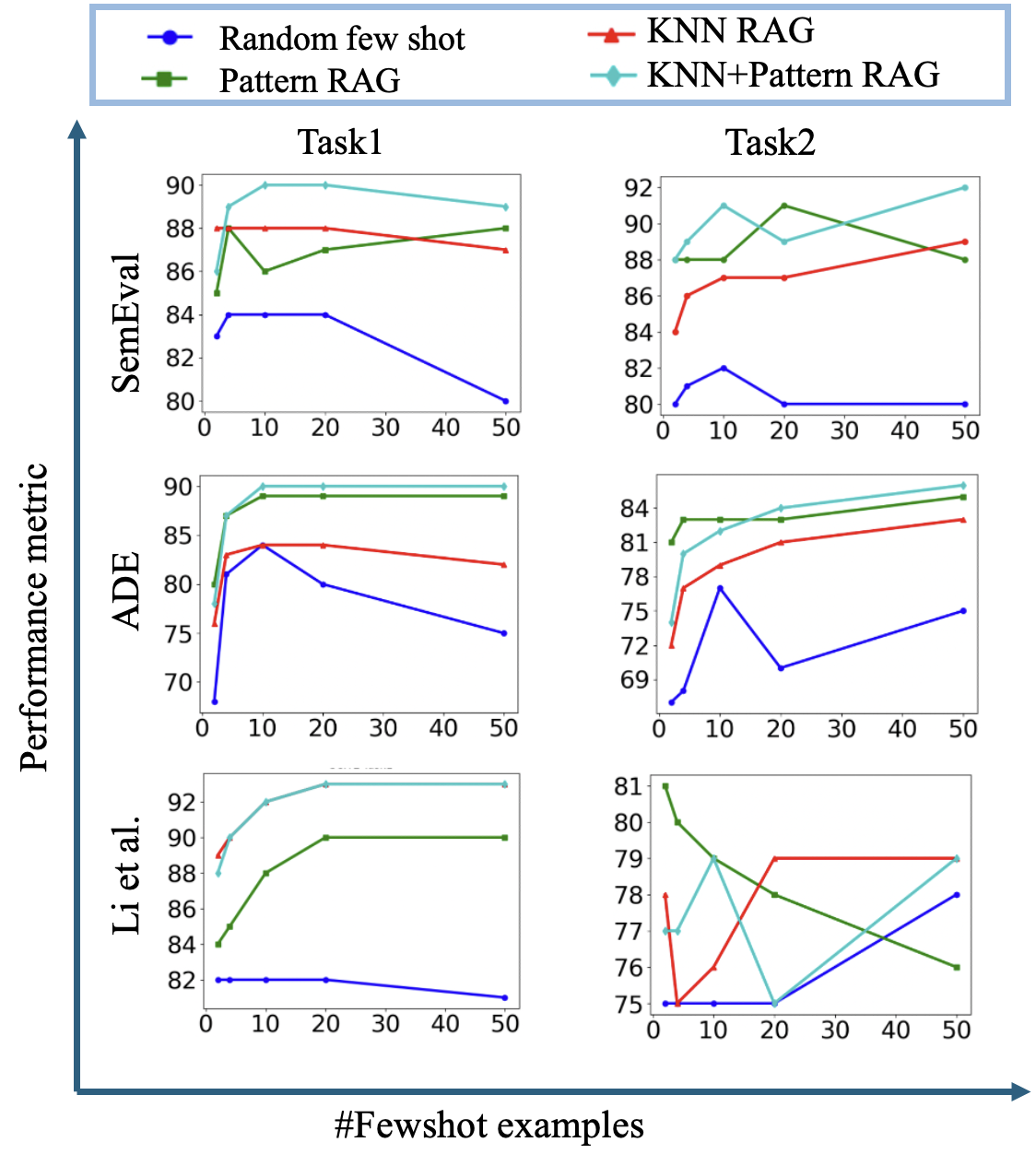}
    \caption{Plot of the number of examples vs. performance results of Causality detection (Task1) and Causality extraction (Task2) from GPT-4o model for different RAG based prompting}
    \label{fig:acc_examplePLOT}
    \vspace{-0.2in}
\end{figure}

\section{Performance vs Fewshot Count} 

In this experiment we validate whether the number of fewshot examples affects
the performance of different promptings: Random fewshot, $k$NN RAG, Pattern RAG,
and $k$NN+Pattern RAG. For obvious reason, Zeroshot is not used as it does not use
any example. We show results for GPT-4o as this is the best performing LLM,
Figure \ref{fig:acc_examplePLOT} shows the results. In this figure, we show six panels.
In each panel, we show the change in F1 score with respect to the number of
fewshot examples for Random fewshot, $k$NN RAG, Pattern RAG and $k$NN+Pattern RAG.
For causality detection, the Random fewshot selection approach shows lower F1
scores than those of all three RAG-based methods, regardless of the number of
examples chosen. This indicates that the causality detection task benefits
solely from the quality of the examples chosen across all three datasets, rather
than the number of examples.

For causality extraction, we see that the RAG-based approaches show higher accuracy than the Random fewshot approach for different fewshot example counts in the SemEval and ADE datasets. In the \citeauthor{li2021causality} plot, we see that the Random fewshot outperforms Pattern RAG at a fewshot example count of 50. This is due to the fact that Pattern RAG dynamically selects examples based on the existence of causal connectives. Our \textit{Fewshot example DB} has a limit of 10 examples for each pattern (causal connectives), so if an instance does not have multiple causal connectives, Pattern RAG cannot obtain more examples. Therefore, when the number of examples is increased to a high value like 50, Pattern RAG does not necessarily has 50 examples, while Random fewshot always have 50 examples.

We conclude from these results that simply increasing the number of examples does not improve LLM performance. The performance increase in the $k$NN+Pattern RAG in most of the LLM results is solely based on the quality of the examples chosen by concatenation, not on the increase in example count.

\begin{table*}
  \small 
  \setlength{\abovecaptionskip}{1.5em}  
\centering
\addtolength{\tabcolsep}{-0.01em}
\begin{tabular}{@{} p{4.5cm} |p{2.5cm}| p{2.5cm}|p{2.3cm}| p{2.5cm} @{}}

\toprule   
   \multirow{2}{*}{\hspace{7.em} Input Sentence} &\multicolumn{2}{c|}{DEPBERT}  & \multicolumn{2}{c}{Pattern RAG GPT 4o} \\
   & \hspace{2.em}Cause &\hspace{2.em} Effect &\hspace{2.em} Cause & \hspace{2.em} Effect \\
 \midrule
  


   \rowcolor{gray!10}  \textcolor{blue}{Heat ,  wind } and  \textcolor{blue}{smoke}  cause  \textcolor{red}{flight delays.}& \textcolor{black}{$\times$ } Heat& \textcolor{black}{\checkmark}flight delays & \textcolor{black}{\checkmark}Heat ,  wind,  smoke & \textcolor{black}{\checkmark} flight delays \\

    Information about  \textcolor{red}{the foodborne illness}  caused by  \textcolor{blue}{salmonella bacteria .} & \textcolor{black}{$\times$ } by  salmonella bacteria &  \textcolor{black}{$\times$ } foodborne illness & \textcolor{black}{\checkmark} salmonella bacteria& \textcolor{black}{$\times$ } foodborne illness \\

   \rowcolor{gray!10}  \textcolor{red}{Eye discomfort}  from  this \textcolor{blue}{staring effect}  is exacerbated by low humidity. & \textcolor{black}{$\times$ }low humidity& \textcolor{black}{\checkmark} Eye discomfort, staring& \textcolor{black}{$\times$ }staring effect, low humidity  & \textcolor{black}{\checkmark} Eye discomfort 
\\

\bottomrule

\end{tabular}
\caption{Comparitative analysis of DEPBERT and the proposed approach on cause-effect extraction tasks. Cause and effect labels are indicated by blue and red color in the input sentence respectively. \textcolor{black}{$\times$} indicates incorrect predictions, while \textcolor{black}{\checkmark} denotes correct ones based on our evaluation metrics.}
\label{tab:multi C-E extract performance}
\end{table*}

\begin{table*}
 \small   
\centering
\addtolength{\tabcolsep}{-0.01em}
\begin{tabular}{@{} l p{8.5cm} >{\centering\arraybackslash} p{3.2cm} >{\centering\arraybackslash} p{3.2cm} @{} }

   \toprule
   Dataset & Input Sentence & Cause label & Effect label\\
    \midrule

   \rowcolor{gray!10} SemEval & Dogs develop a fever from stress and/or pain such as in a severe flea infestation. &stress & fever\\

   ADE & Benzocaine-induced methemoglobinemia has been reported in man, dogs, and cats. & Benzocaine & methemoglobinemia\\

   \rowcolor{gray!10} \citeauthor{li2021causality} &Paralysis  or  convulsions  are caused by  hormone deficiencies and imbalances .& \textbf{--} hormone deficiencies and imbalances &Paralysis\\
   \rowcolor{gray!10}      &      & \textbf{--} hormone deficiencies and imbalances& convulsions\\

  \bottomrule

\end{tabular}
\caption{Input examples for causality extraction. }
\label{tab:Task2: Cause-Effect phrase extraction input examples}
\vspace{-0.2in}
\end{table*}

\section{Causality extraction on multi-word and multi-cause-effect scenarios.} 
The dataset by \citeauthor{li2021causality} contains sentences labeled with multiple cause-effect pairs. Table \ref{tab:multi C-E extract performance} presents a comparative analysis of the transformer-based DEPBERT method and our best-performing Pattern RAG method using GPT-4o. Extracting multiple cause-effect pairs from a sentence poses significant challenges for traditional deep learning methods. However, our proposed LLM-based RAG method demonstrates a notable improvement in performance (F1 score). From the first example, we observe that DEPBERT struggles to predict all causes and effects correctly in a multi-labeled dataset. Our approach shows a good performance improvement but is hindered by the exact-match  requirement of the evaluation metric (all words in the label should be present in the predicted phrase). For instance, in example two, the actual effect phrase is "the foodborne illness," while our model predicts "foodborne illness," resulting in a misclassification due to the missing article. Example three highlights a scenario where our model predicts the cause phrase correctly, but the dataset lacks corresponding labels. This analysis shows that the causality extraction for a  multi-word and multi-cause-effect scenarios, can be improved by a better evaluation metrics and labeling practices.

\section{Conclusion}

In this work, we proposed two retrieval-augmented generation (RAG) based dynamic prompting methods for LLMs to address the task of causality mining.
These approaches leverage both semantic and pattern-based similarity between the input sentence and a collection of test sentences to construct dynamic prompts. Experimental results demonstrate that our proposed methods significantly enhance the performance of causality detection and causality extraction tasks. 




\section{Limitations} 
One major limitation of our approach is its focus on intra-sentential causality relations, without addressing inter-sentential causality relations. Extending our method to detect and extract inter-sentential causality could be a promising direction for future research, offering significant benefits to real-world situations. For example, in the medical domain, understanding complex causal relationships across sentences is crucial.
Besides, causality relations exist across all languages. Our experiments have been conducted only in English language. An in-depth understanding of the meaning and structure of different languages would help extend our RAG-based approaches to causality mining in different languages. 
\section{Acknowledgments}
Dr. Hasan's research is supported by National Science Foundation (NSF) grant number
2417275.

\bibliography{CausalityRAG}

\section{Appendix}

\appendix


The prompts for causal connective extraction, causality detection (zeroshot setting), and causality extraction (zeroshot setting) are shown in Figures \ref{fig:Causal Connective Extraction Prompt}, \ref{fig:Causality Detection Prompt}, and \ref{fig:Causality Extraction Prompt}, respectively.
For different RAG settings and for different causality tasks, the following texts and dynamically generated examples (as retrieved by various RAG methods) are appended to the associated prompt. 
``Below are 10 example sentences which show causality relation  with  their cause and effect 
          phrase delimited between the XML tags <cause>, </cause> , <effect> and </effect>." + EXAMPLES.''



\begin{figure*}
    
\setlength{\abovecaptionskip}{0.1em} 
\setlength{\belowcaptionskip}{2em}   

    \centering
    \includegraphics[width=6.5in]{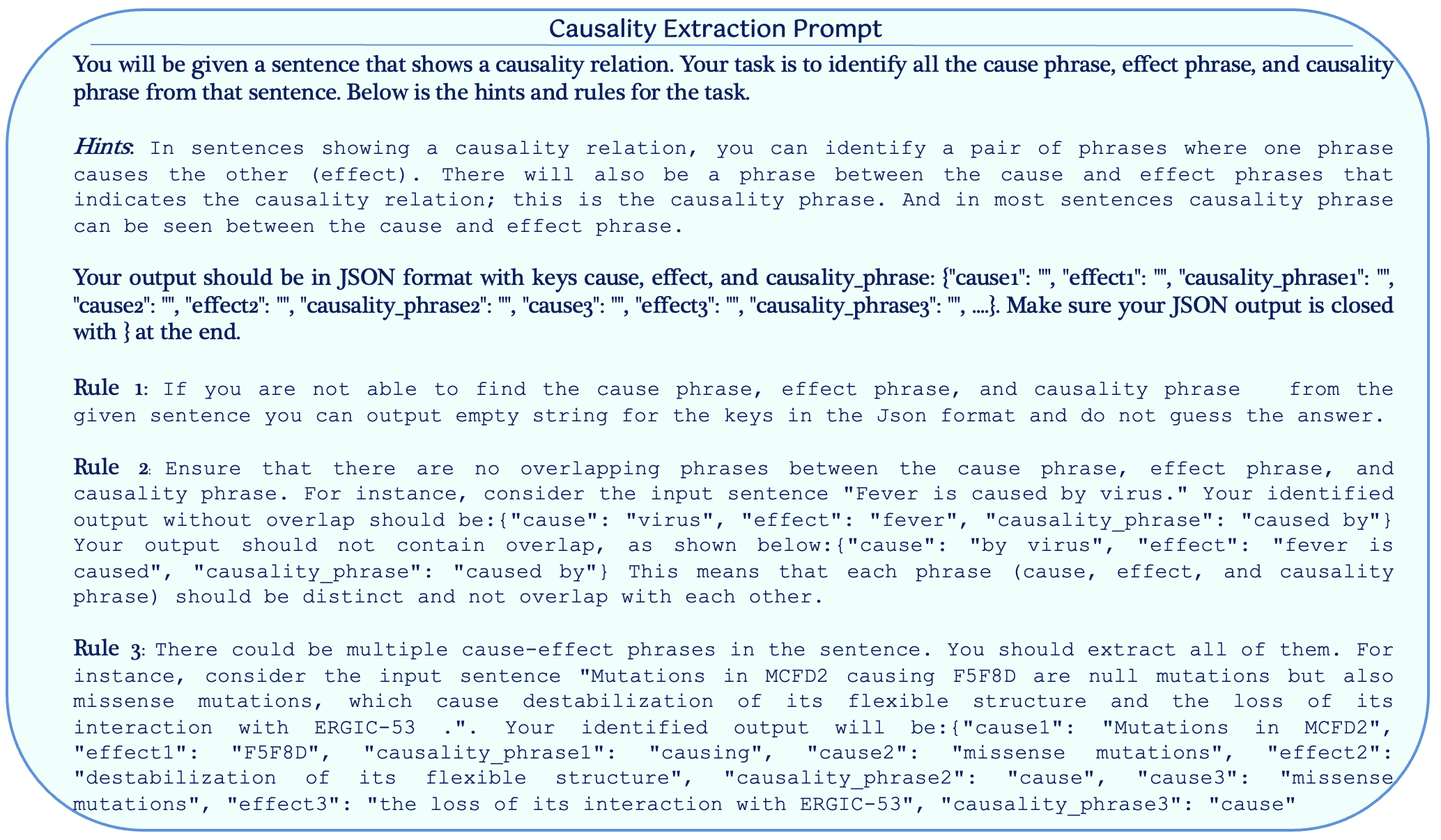}
    \caption{Causality Extraction Prompt
}
    \label{fig:Causality_Extraction_Prompt}
    \vspace{-0.2in}
\end{figure*}

\section{Dataset Discription}
\textbf{SemEval:}
For our causality detection task, we select 249 causal sentences from the test set that are absent in our \textit{Fewshot Example DB} and a random sample of 300 non-causal sentences from the test data (\ref{tab:Dataset Statistics}). We label the causal sentences with a 1 and the non-causal sentences with a 0. Table \ref{tab:Task1 Causality detection input examples} shows a sample of input from the SemEval data for the causality detection task.
For the causality extraction task, we use the same 249 causal sentences from the detection task and their provided cause and effect labels as test keys. Table \ref{tab:Task2: Cause-Effect phrase extraction input examples} shows a sample of input sentences along with their provided cause and effect labels.

\textbf{ADE: }
This dataset contains 6,821 causal sentences involving drugs and their effects. After filtering sentences 
having multiple cause-effect phrases, we retain 4,271 instances. We randomly partition these into 2,000 training instances (later used for the \textit{Fewshot Example DB} creation) and 2,271 test instances. 
For our causality detection task, we use these 2,271 test instances (causal sentences absent in our \textit{Fewshot Example DB}) and combine them with a random sample of 2,000 non-causal sentences from the training dataset of the \citeauthor{li2021causality} dataset. An ADE input sample is shown in Table \ref{tab:Task1 Causality detection input examples}.
For the causality extraction task, we use the 2,271 causal sentences as input and use the drug and effect labels as ground truth.

\textbf{\citeauthor{li2021causality} : }
For the causality detection task,  we use the entire test data (Table \ref{tab:Dataset Statistics}) as our input (all causal sentences are absent in our \textit{Fewshot Example DB}).
For the causality extraction task,  we use the 191 causal sentences as input and use the provided cause-effect labels as our test keys. The \citeauthor{li2021causality} dataset has multiple cause-effect pairs labeled in a given sentence, resulting in a total of 296 unique triplets of (cause, effect, sentence).

\begin{table}
 \small 
 \setlength{\belowcaptionskip}{1.5em}  
\centering
\addtolength{\tabcolsep}{-0.01em}
\begin{tabular}{@{} l p{4cm} >{\centering\arraybackslash}p{1cm} @{} }
   \toprule
   Dataset & Input Sentence & label \\
    \midrule

   \rowcolor{gray!10} SemEval & Dogs develop a fever from stress and/or pain such as in a severe flea infestation. & 1\\
   \rowcolor{gray!10} &The researchers placed the compound in a tube, which then was put inside a magnet. & 0\\

   ADE &Benzocaine-induced methemoglobinemia has been reported in man, dogs, and cats. & 1\\
   &The man placed the cartridge into the printer. &0\\

   \rowcolor{gray!10} \citeauthor{li2021causality} &Paralysis  or  convulsions  are caused by  hormone deficiencies and imbalances .. & 1\\
  \rowcolor{gray!10}  & This  theme  has been covered in  science fiction  like Star Trek. &0\\
   
\bottomrule

\end{tabular}
\caption{Task1 Causality detection input examples }
\label{tab:Task1 Causality detection input examples}
\vspace{-0.2in}
\end{table}

\begin{figure*}
 \small   
\setlength{\abovecaptionskip}{0.1em} %
\setlength{\belowcaptionskip}{1.5em}  
    \centering
    \includegraphics[width=3.4in]{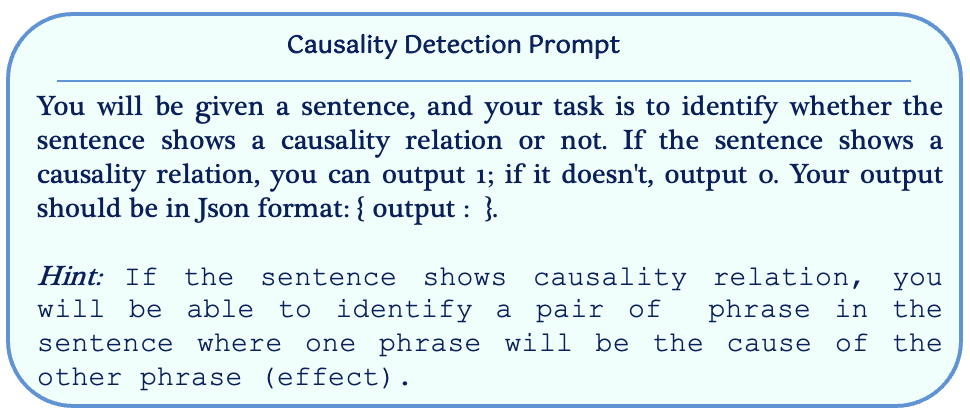}
    \caption{Causality Detection Prompt
}
    \label{fig:Causality Detection Prompt}
    \vspace{-0.2in}
\end{figure*}

\begin{figure*}
 \small   
\setlength{\abovecaptionskip}{0.1em} 
\setlength{\belowcaptionskip}{1.5em}   
 
    \centering
    \includegraphics[width=3.4in]{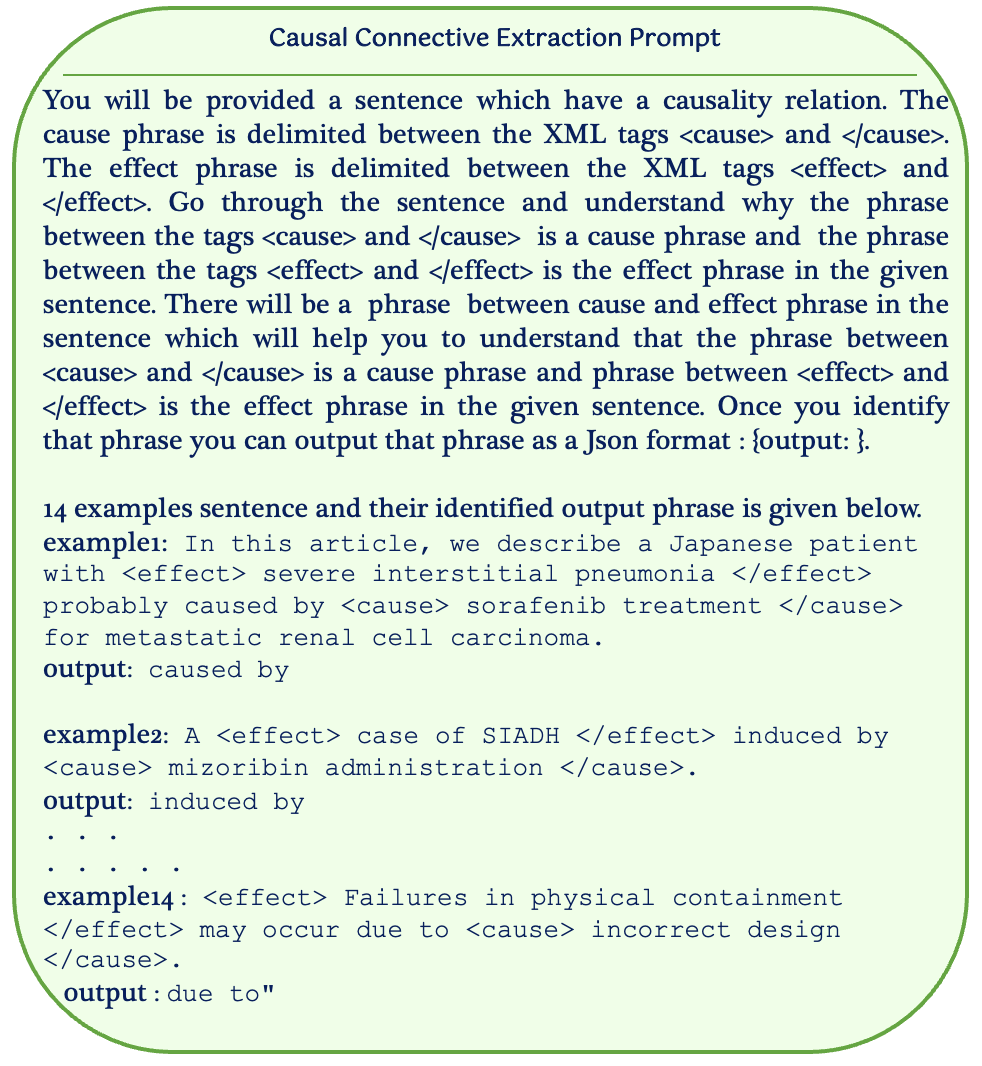}
    \caption{Causal Connective Extraction Prompt}
    \label{fig:Causal Connective Extraction Prompt}
    \vspace{-0.2in}
\end{figure*}




\begin{table*}
\small    
\centering
\addtolength{\tabcolsep}{-0.01em}
\begin{tabular}{@{}p{1.5cm} p{1.5cm} p{1.2cm} p{1.2cm} p{1.7cm} @{}}
   \toprule
  Dataset & Split & \#total unique instances &\#causal relations  & \#non-causal relations \\
   \midrule
 \rowcolor{gray!10} \textit{Fewshot Example DB} &\multirow{2}{*}{-} & \multirow{2}{*}{2,365} & \multirow{2}{*}{2,365} & \multirow{2}{*}{-}\\
   
  \multirow{4}{*}{SemEval} & Train& 8,000 &1,003  & 6,997 \\
    & Test & 2,717 &328 & 2,389 \\
    &Task1 input &549 & 249& 300\\
    & Task2 input &249 &249 & -\\
   
  \rowcolor{gray!10}  & Train& 2,000 &2,000  & - \\
  \rowcolor{gray!10} ADE & Test & 2,271  &2,271  & - \\
  \rowcolor{gray!10}  &Task1 input &4,271 & 2,271& 2,000\\
  \rowcolor{gray!10}  & Task2 input &2,271 &2,271 & -\\

  \multirow{4}{*}{\citeauthor{li2021causality}} & Train& 4,450 & 1,079 & 3,371 \\
    & Test & 786 &191  & 595 \\
    &Task1 input & 786 &191 & 595\\
    & Task2 input & 191&191 & -\\
    
  \bottomrule

\end{tabular}
\caption{Dataset statistics. (Task1: Causality detection, Task2: Causality extraction) }
\label{tab:Dataset Statistics}
\vspace{-0.2in}
\end{table*}


\end{document}